\documentclass[letterpaper, 10 pt, conference]{ieeeconf}
\IEEEoverridecommandlockouts
\usepackage{multirow}
\usepackage{cite}
\usepackage{amsmath,amssymb,amsfonts}
\usepackage{graphicx}
\usepackage{textcomp}
\usepackage{tabularx}
\usepackage{amsmath}
\usepackage{dblfloatfix}
\usepackage{siunitx}
\usepackage{booktabs}
\usepackage{longtable}
\usepackage{float}
\usepackage{gensymb}
\usepackage{balance}

\title{\LARGE \bf HyperSurf: Quadruped Robot Leg Capable of Surface Recognition with GRU and Real-to-Sim Transferring

\author{Sergei Satsevich, Yaroslav Savotin, Danil Belov, Elizaveta Pestova, Artem Erhov, Batyr Khabibullin, \\
Artem Bazhenov, Vyacheslav Kovalev, Aleksey Fedoseev and Dzmitry Tsetserukou}

\thanks{The authors are with the Intelligent Space Robotics Laboratory, CDE, Skolkovo Institute of Science and Technology, Bolshoy Boulevard 30, bld. 1, 121205, Moscow, Russia
Email: {\tt\small \{Sergei.Satsevich, Yaroslav.Savotin, Danil.Belov, Elizaveta.Pestova, Artem.Erkhov, Batyr.Khabibullin, Artem.Bazhenov, Vyacheslav.Kovalev, Aleksey.Fedoseev, Dzmitry.Tsetserukou\}@skoltech.ru\ }
  }
}
\begin{document}
\maketitle
\thispagestyle{empty}
\pagestyle{empty}

\begin{abstract}
This paper introduces a system of data collection acceleration and real-to-sim transferring for surface recognition on a quadruped robot. The system features a mechanical single-leg setup capable of stepping on various easily interchangeable surfaces. Additionally, it incorporates a GRU-based Surface Recognition System, inspired by the system detailed in the DogSurf paper \cite{DogSurf}. This setup facilitates the expansion of dataset collection for model training, enabling data acquisition from hard-to-reach surfaces in laboratory conditions. Furthermore, it opens avenues for transferring surface properties from reality to simulation, thereby allowing the training of optimal gaits for legged robots in simulation environments using a pre-prepared library of digital twins of surfaces. Moreover, enhancements have been made to the GRU-based Surface Recognition System, allowing for the integration of data from both the quadruped robot and the single-leg setup. The dataset and code have been made publicly available. 
\end{abstract}
\noindent

\textbf{\textit{Keywords: Robotics, Quadruped Robot, Surface Recognition, Terrain classification, IMU, GRU}}




\maketitle

\section{Introduction}

Quadruped robots get a lot of attention nowadays, due to their potential advantage over wheeled robots to navigate unstructured off-road terrains and urban environments where people live and work. This ability opens a lot of potential applications in different spheres like autonomous inspection, agriculture \cite{agricultural}, delivery \cite{alphred}, assisting for visually impaired people, space exploration. However, performing some tasks demands perception of different terrains and consequent adeptness to them or awareness of the user about the type of a surface. 

Employing quadruped robots as assistants requires an effective system, which is able to evaluate whether a traversed surface is dangerous or not in urban environments. One of the crucial criteria in this case is the robot's ability to evaluate surface slipperiness. A slippery surface significantly increases the risk of severe injuries, e.g., hip and spine cord fractures from falls. The statistic estimations suggest that hip fracture rates will nearly double by 2050 compared to 2023 levels \cite{Sing_CW_bones}. Consequently, an assistant quadruped robot must possess the advanced intelligence to warn people about slippery surfaces.
\begin{figure}[hbt]
    \centering
    \includegraphics[width=1\linewidth]{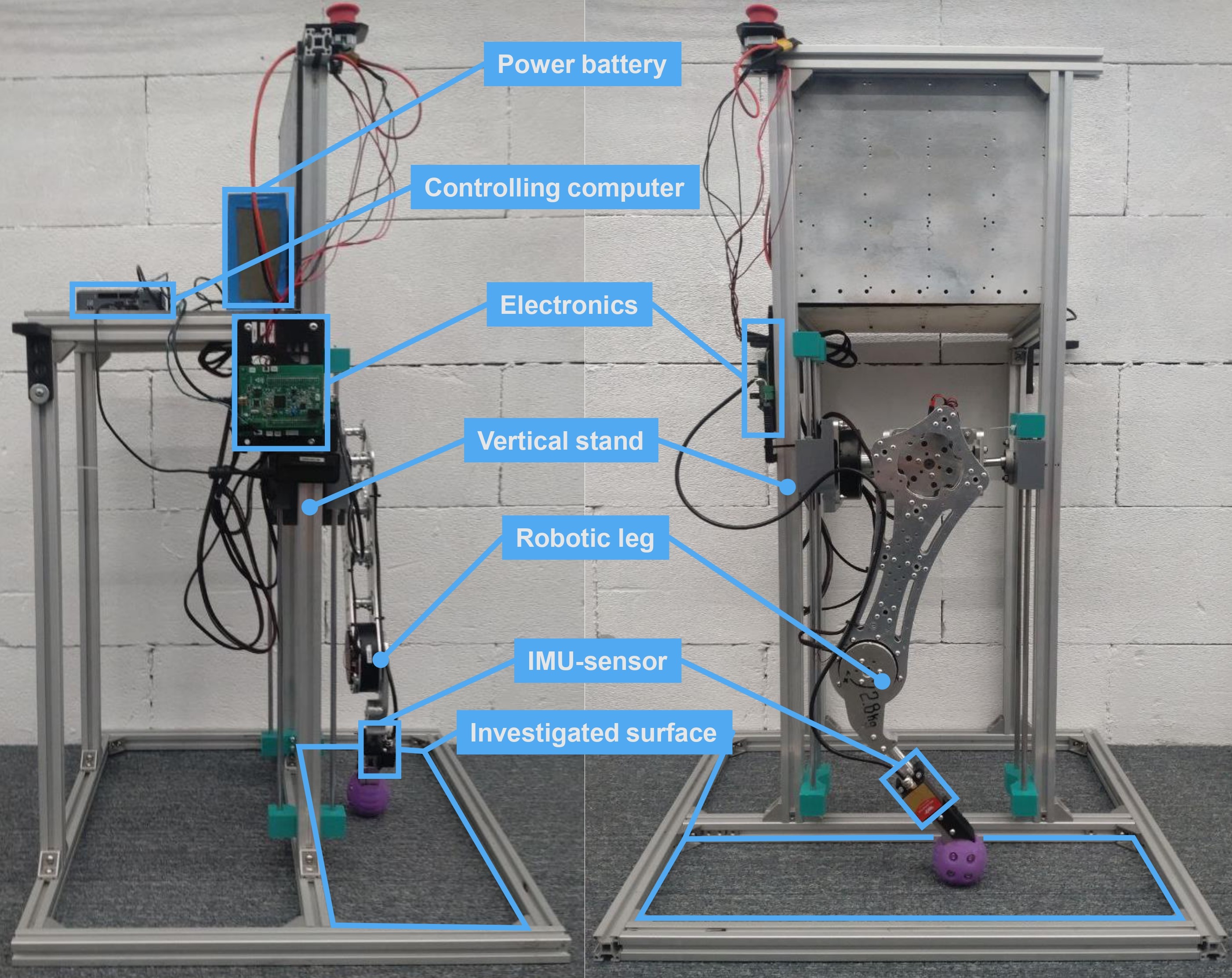}
    \caption{The experimental stand consists of a leg, fixing stand, control electronics, IMU sensor, and the surface sample under investigation.}
    \label{fig:Stand_overview}
\end{figure}
Investigating the surfaces of celestial bodies involves traversing regolith, which is the predominant material covering the surfaces of Mars, the Moon, and various other celestial bodies. The regolith is often loose, granular, and heterogeneous, posing challenges for traversal with wheeled robots. \cite{kolvenbach2021traversing}. Therefore, using quadruped robots to explore the complex surfaces of celestial bodies is rational due to their ability to function and traverse intricate landscapes. However, locomotion in complex environment demands from legged robots to adapt their behavior to different parameters of surfaces, which stimulate to develop different approaches for surface recognition.  

Our method builds upon the DogSurf \cite{DogSurf} surface recognition system and expands the effectiveness of its method. In this paper, we have shifted our approach to dataset collection from gathering data by walking on a quadruped robot to collecting data in a laboratory setting while the robot stands on a single leg. This method offers advantages in terms of speed, simplicity, and cost-effectiveness, enabling faster data collection and experimentation, because our new one-leg setup does not require investment of manpower and time for datasets collections. Our GRU-based system offers a cost-effective solution that can be easily implemented across a wide range of legged robotics. Utilizing this setup would facilitate the collection of datasets, which could then be transferred to a quadruped robot or in simulation environment allowing to develop optimal locomotion patterns in simulation. By incorporating feedback from IMU sensor systems and adjusting parameters, this approach can enhance the adaptability of legged robots. 

In this work, we have expanded our research by systematizing the creation of digital surface duplicates within the Gazebo simulation environment, exemplified through the replication of a floor tile surface. The current state of simulation environments commonly used by the robotics community, such as Gazebo and NVIDIA’s Isaac, lacks extensive libraries of pre-defined surface types that accurately reflect the diversity of real-world terrains. This deficiency presents a significant challenge during the Sim-to-Real deployment phase, where discrepancies between simulated and actual surfaces often lead to substantial errors in robot behavior \cite{tan2018sim}. Walking robots, particularly quadrupeds, face difficulties when transitioning from controlled simulation environments to the complexities of real-world surfaces, which can differ greatly from the standard options available in these simulations. Through the development and implementation of these digital surface duplicates, we aim to reduce the trial-and-error phase associated with surface recognition and locomotion in physical environments, thereby streamlining the path from simulation to real-world deployment.

\section{Related Works}


DogSurf is a novel method employing quadruped robots to aid visually impaired individuals in real-world navigation. This approach empowers the quadruped robot to identify slippery surfaces and provide auditory and tactile cues to signal the user when to halt. Additionally, a cutting-edge GRU-based neural network architecture boasting a mean accuracy of 99.925\% was introduced for the multiclass surface classification task tailored specifically for quadruped robots. The dataset used for training was gathered using a Unitree Go1 Edu robot.

In addition to DogSurf, several other methodologies for surface recognition are available or currently employed in quadruped robot locomotion.


\textbf{First approach}  involves employing various tactile estimation methods. Weerakkodi et al. \cite{Nipun} introduced the utilization of Touch Sensitive Foot (TSF) tactile sensors, positioned on the dog’s feet to gather data, subsequently processed by Convolutional Neural Networks (CNN). This method enables the attainment of a validation accuracy of 74.37\% and a peak recognition score of 90\% for line patterns. Despite its potential, this approach faces several limitations, such as the high service cost due to direct sensor contact with the surface, leading to gradual wear and tear over time.

\textbf{Second approach} involves utilizing audio data. Dimiccoli et al. \cite{Dimiccoli} employed a gripper to manipulate objects and captured audio signals, subsequently utilized for CNN training, resulting in approximately 85\% accuracy. Another successful attempt was made by Vangen et al. \cite{Vangen}, who suggested employing a sensorized paw for audio-based terrain classification, achieving around 78\% accuracy. However, this method exhibits limited real-world applicability due to the diverse ambient sounds present, which notably degrade the neural network's accuracy.

\textbf{Third approach} involves utilizing Force/Torque (F/T) sensor data either independently or in conjunction with the aforementioned methods. Bednarek et al. \cite{Bednarek_F/t} advocated for the utilization of data solely from F/T sensors on the quadrupedal robot ANYmal \cite{ANYmal}. Additionally, they introduced a novel CNN-1d convolution architecture and devised a clustering method for terrain classification, achieving an accuracy of 93\%. Another method employing a Transformer architecture was proposed in \cite{HAPTR2} and attained an accuracy of 97.33\% on a QCAT dataset. Kolvenbach et al. \cite{Kolvenbach_f/c_2} recommended combining F/T and IMU sensors for the ANYmal robot and developed a specialized test stand, achieving an accuracy of 98.6\%.

On average, the IMU-based approach yields superior results and offers advantages such as no direct environmental contact, resilience to weather and time-of-day variations, and cost-effectiveness. This method has been successfully implemented in a walking robot utilizing gated recurrent units (GRU) \cite{GRU}.


\section{System Overview}
\subsection{Hardware Architecture}
To conduct experiments to determine surface types, a special experimental setup was created to ensure repeatability and uniformity of the data obtained when interacting with various surface types. The creation of the stand significantly simplified and reduced the cost of the experiment, as it eliminated the need to use a full-fledged quadruped robot. The main components of the stand (\ref{fig:Stand_overview}) include a full-size leg of a quadruped robot with three degrees of freedom, mounted on a vertical stand, control electronics, an IMU sensor, and the surface for investigation.

\begin{figure}[hbt]
    \centering
    \includegraphics[width=1\linewidth]{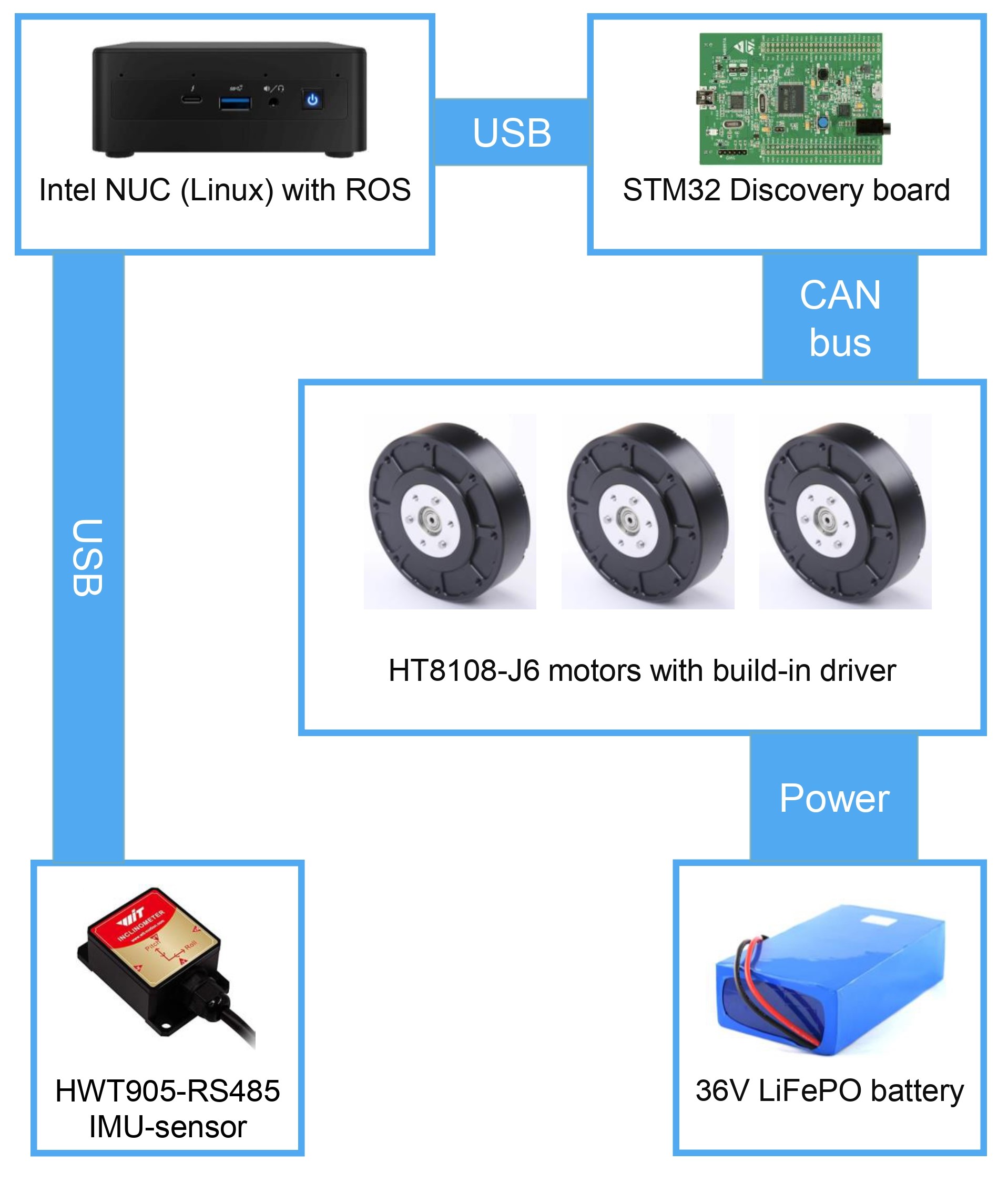}
    \caption{HyperSurf system overview, including modules and interfaces developed for surface recognition.}
    \label{fig:System_overview.jpg}
\end{figure}

The leg mounting system is designed to allow the leg to move smoothly along the vertical guides, ensuring both stability and consistent, repeatable movements. The vertical guide mechanism is essential for controlling the range of motion, preventing lateral deviations, and ensuring that each motion, particularly the jumps, follows the intended trajectory. The leg's movements are regulated by the electronic control unit, which sends commands to coordinate the bending and extension of the joints. This coordinated action enables the leg to perform jumping motions, lifting off the surface by approximately 20 mm. These jumps are vital for allowing the robot to interact with the surface under investigation, gathering information without excessive force or disruption. At the end of the leg, an Inertial Measurement Unit (IMU) sensor, specifically the HWT905-RS485 model, is attached. This sensor plays a crucial role in the feedback loop by providing real-time data on the leg's position, orientation, and movement. The IMU allows the system to monitor the performance of each jump. 

The leg was designed and manufactured as part of a project focused on developing an autonomous quadruped robotic system. It consists of 3 movable joints and 3 BLDC electric motors that drive its movement. The joint in contact with the surface is equipped with a resilient silicone tip to cushion the contact.

The control of electric motors is performed using an algorithm in the micro-ROS environment on the Intel NUC microcomputer. This same computer is responsible for collecting data from the IMU sensor. The entire system is designed for reliable and stable reproduction of the jumping pattern specified by the user on the surface under investigation.

\subsection{HyperSurf Control System}

In the control system of the HyperSurf robotic leg, a combination of ROS2 and micro-ROS technologies governing its movements. Within the ROS2 package, specific leg trajectories are defined within Python scrips encapsulated in ROS2 packages. These trajectories, created to facilitate leg movement, are translated into motor angles, shaping the physical actions of the robotic limb. The computations required for these calculations were executed on an Intel NUC computer. 

Once the trajectory calculations are finalized, they are converted into motor angles by an inverse kinematics algorithm. These angles, along with velocity and the coefficients $K_p$ - desired position stiffness and $K_d$ - velocity gain, are transmitted to the leg actuators via the /Hyper dog joints/Commands topic. The stiffness and damping coefficients were selected in advance, taking into account the load on each motor. They determine how quickly the desired position will be achieved and how rigidly this position will be held when external forces are applied. Sending data to the motors is feasible due to the created custom ROS2 message. The data is transmitted in three separate messages due to the leg's configuration with three motors, each motor possessing a unique identifier (can\_id). This enables the distinction between signals sent over a single CAN bus.

The micro-ROS agent facilitates the seamless interaction between the high-level computation environment and the low-level hardware control. This agent establishes a vital link between the ROS2 ecosystem running on the computer and the microcontroller system managing the hardware. The /joints/Commands topic published by the ROS2 environment is subscribed to by the /STM\_node. The micro-ROS agent processes these commands and translates them into Pulse Width Modulation (PWM) signals, which are then sent to the BLDC motor drivers. By facilitating this communication, the micro-ROS agent ensures that commands from the ROS2 system are accurately and efficiently executed by the hardware, allowing for precise control of the robotic system’s movements and operations.

HyperSurf not only executes the prescribed leg trajectories but also interfaces with motor encoders, capturing critical feedback on the limb's position and velocity. This encoder data is meticulously processed, providing insights into the leg's current state and ensuring precise control over its motion.

\subsection{Conducting the Experiment}

To verify the effectiveness of surface recognition using feedback from the IMU sensor, a diverse set of surfaces with distinct properties was assembled for testing. The robotic leg's interaction was assessed on four different types of surfaces: an EVA polymer sheet (A), a textured ceramic surface (B), a smooth enamel-coated ceramic tile (C), and a carpet (D).

The setup ensured that the leg stand remained stationary to avoid any vibrations that might interfere with the performance of the neural network. One of the surfaces was then placed underneath the robotic leg, ensuring it stayed immobile. These jumping tests lasted between 5 to 7 minutes on each surface to gather sufficient data, repeating this process four times for each surface. The surfaces were changed and the jump frequency was varied, ranging from one jump every 0.8 - 2 seconds. Data from the IMU sensor was recorded using its API throughout the experiment. This data was subsequently processed and transformed to prepare it for the training of the neural network, ensuring that the model could accurately recognize and differentiate between the various surfaces based on the sensor feedback.

\subsection{Dataset Structure}
Throughout the study, a dataset comprising over 1,000,000 samples (equivalent to approximately 1.5 hours of data) was amassed. Data collection encompassed four classes: a sheet of EVA polymer, a textured ceramic surface, a smooth enamel-coated ceramic tile surface, and carpet. During the collection process, the stand operated in automatic stepping mode, varying its frequency to ensure an adequate diversity of data.  This variation in frequency was crucial for ensuring that the dataset captured a wide range of interaction scenarios, providing the necessary diversity to improve the robustness and generalization of the subsequent analysis. The large volume of data collected from these different surfaces was critical for training and refining the surface recognition algorithms used in the study.


\subsection{GRU-based Surface Recognition}
Our previous method, DogSurf module utilizes a GRU neural network as the foundation. The IMU data from the accelerometer and gyroscope is collected and appended to a queue containing 100 samples. During each iteration, a Forward Pass is conducted through the first \textbf{Standard Scaler}, then through \textbf{Principal Component Analysis (PCA) transform}, and then through \textbf{bidirectional GRU} neural network. 

\begin{figure}[hbt]
    \centering
    \includegraphics[width=0.96\linewidth=8cm]{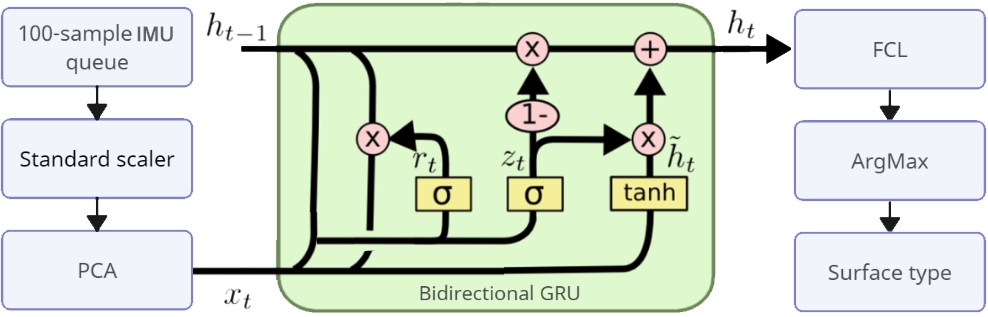}
    \caption{Neural network architecture.}
    \label{fig:Neural network architecture.}
\end{figure}

 \textbf {Code and other supplementary materials can be accessed by link: https://github.com/Alfasaz/HyperSurf. \cite{HyperSurf.git}} The results of DogSurf module performance of the HyperSurf compared to the State-of-the-Art are listed in Table \ref{tab:results}.

\begin{table}[htp]
\centering
\caption{Comparison of HyperSurf with the State-of-the-art Approaches.}
\label{tab:results}
\begin{tabular}{c|c|c|}
  \cline{2-3}
  & Model & Accuracy\\
  \hline
   \multicolumn{1}{|c|}{Tactile} & Weerakkodi et al. \cite{Nipun} & 0.74370\\
   \hline
   \multicolumn{1}{|c|}{\multirow{2}{*}{Audio}} & Dimiccoli et al. \cite{Dimiccoli} & 0.84700\\
   \cline{2-3}
   \multicolumn{1}{|c|}{}& Vangen et al. \cite{Vangen} & 0.77900\\
   \hline
   \multicolumn{1}{|c|}{\multirow{2}{*}{F/T}} & HAPTR2 \cite{HAPTR2} & 0.97330\\
   \cline{2-3}
   \multicolumn{1}{|c|}{} & Jakub Bednarek et al.  \cite{Bednarek_F/t} & 0.93590\\
   \hline
   \multicolumn{1}{|c|}{F/T \& IMU} & Kolvenbach et al. \cite{Kolvenbach_f/c_2} & 0.98600\\
   \hline
   \multicolumn{1}{|c|}{\multirow{3}{*}{IMU}} & Lomio et al. \cite{lomio} & 0.64950\\
   \cline{2-3}
   \multicolumn{1}{|c|}{}& Singh et al. \cite{singh} & 0.88000\\
   \cline{2-3}
   \multicolumn{1}{|c|}{} & \textbf{DogSurf} & \textbf{0.99925}\\
  \hline
\end{tabular}
\end{table}

\begin{figure}[hbt]
    \centering
    \includegraphics[width=8 cm]{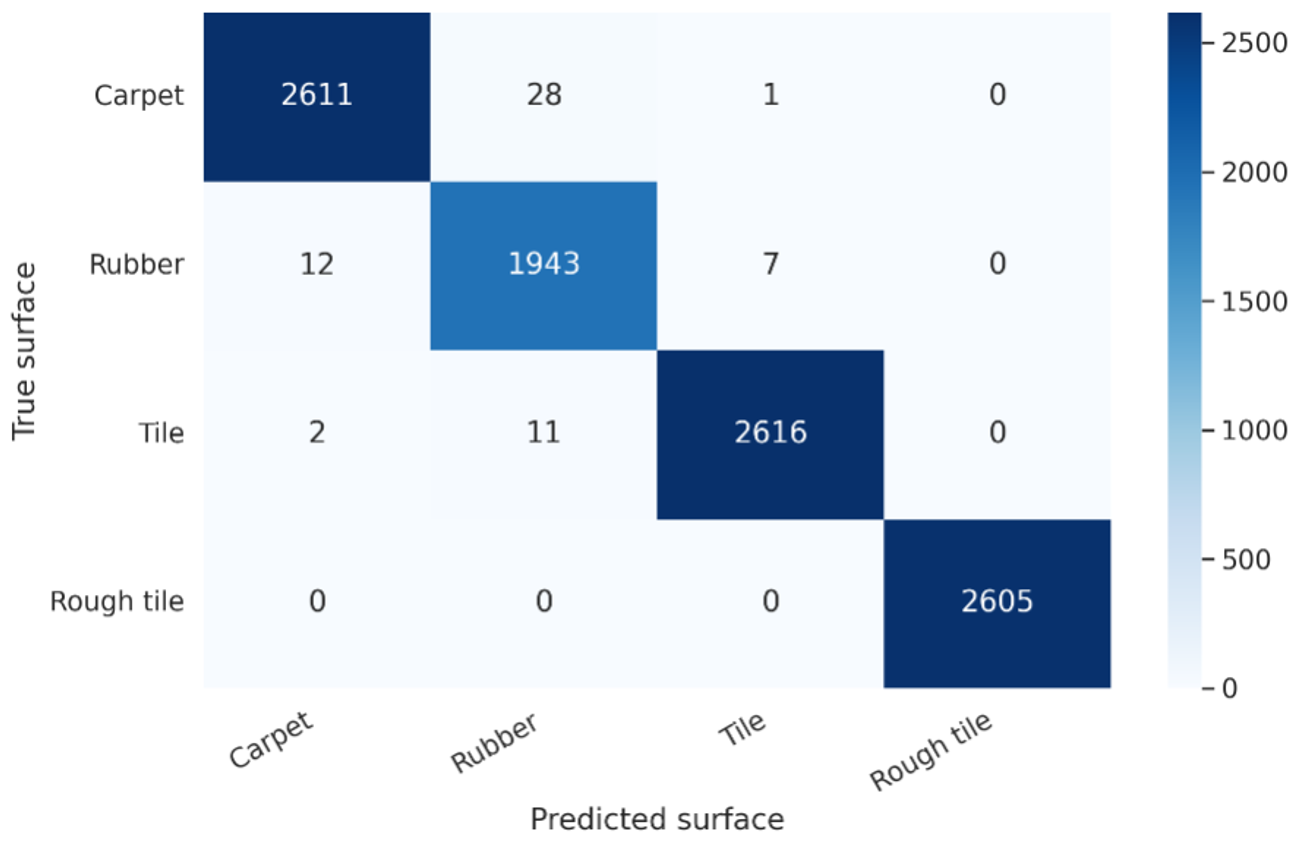}
    \caption{Confusion matrix for HyperSurf.}
    \label{fig:conf2.png}
\end{figure}

For HyperSerf we get the new results which represented in Table \ref{tab:results-surf} and confusion matrix in Fig. \ref{fig:conf2.png}.

\begin{table}[htp]
\centering
\caption{Surface Recognition by HyperSurf}
\label{tab:results-surf}
\begin{tabular}{c|c|c|c|}
  \cline{2-4}
  & Precision & Recall & F1 score\\
  \hline
  \multicolumn{1}{|c|}{Carpet} & 0.99467 & 0.98902 & 0.99183\\
  \hline
  \multicolumn{1}{|c|}{Rubber} & 0.98032 & 0.99032 & 0.98529\\
  \hline
  \multicolumn{1}{|c|}{Tile} & 0.99695 & 0.99506 & 0.99600\\
  \hline
  \multicolumn{1}{|c|}{Rough tile} & 1.00000 & 1.00000 & 1.00000\\
  \hline
\end{tabular}
\end{table}

The outstanding results of surface recognition open new perspectives of utilizing quadruped robots both as companions, and also in complex landscapes.

\section{Creating a Digital Twin of HyperSurf System}

To conduct the stand simulation in the Gazebo environment, a digital twin was created based on the following steps:

\subsection{Preparation of the stand description file}

The three-dimensional model of the stand was meticulously crafted within Computer-Aided Design (CAD) software, leveraging a simplified rendition of the complete quadruped robot designed in Siemens NX. This comprehensive model encompasses all integral components essential for the stand's functionality.

To seamlessly integrate the CAD model into the Gazebo simulator, a Unified Robot Description Format (URDF) was meticulously generated. Crafted through Fusion 360, this URDF file intricately delineates all stand components, their interactions, and kinematic properties. Notably, the URDF generation process was facilitated by a Python script available at the repository \cite{fusion2urdf}.

In line with the real-world stand setup, an IMU sensor was seamlessly incorporated into the dog's leg using ROS tools. This addition faithfully mirrors the physical stand's configuration, thereby enhancing the fidelity and realism of the simulation environment.

Furthermore, to accurately simulate various surfaces such as tiles, rubber, and brick within the Gazebo environment, meticulous specifications were established. These surface characteristics were detailed comprehensively in the URDF file, with each parameter meticulously tuned to replicate the physical attributes of the real-world surfaces. Key parameters, including the friction coefficient, damping coefficient, and stiffness coefficient, were precisely calibrated to ensure that the simulation dynamics closely mirror the actual conditions. The process of transferring data on surface properties from the real world to the simulation involved careful calibration of these coefficients. The goal was to align the virtual IMU data collected during simulated experiments (leg jumps) with the IMU data recorded in real-world tests. This iterative calibration process ensures that the virtual environment accurately reflects the behavior of the physical surfaces, allowing for a realistic and reliable simulation of surface interactions.

\subsection{Kinematics Control Implementation}

Leg kinematics control was realized using symbolic and numerical computation techniques. Forward kinematics determined leg component positions, while inverse kinematics was used to adjust the joint angles to achieve specific end-effector positions. Symbolic computation was performed using SymPy, which facilitated the calculation of transformation matrices and coordinates essential for understanding the geometric relationships between different parts of the leg. CasADi was utilized for numerical optimization, handling the task of adjusting joint angles to meet the required end-effector positions with high precision. This integrated approach, combining symbolic and numerical methods, enabled fine-tuned control of the leg's movements. It ensured effective interaction with various surfaces and environments within the Gazebo simulator, allowing for robust data collection and analysis.

\subsection{Dataset Collection}

In the Gazebo simulator, utilizing the digital twin of the stand, a dataset was gathered to train the surface recognition model. Similar to the real stand, data from the IMU sensor was collected and analyzed. This process aimed to demonstrate the transfer ability of data between the physical and simulated environments, showcasing the bidirectional capability of real-to-simulation and simulation-to-real data integration.

\begin{figure}
    \centering
    \includegraphics[width=1.0\linewidth]{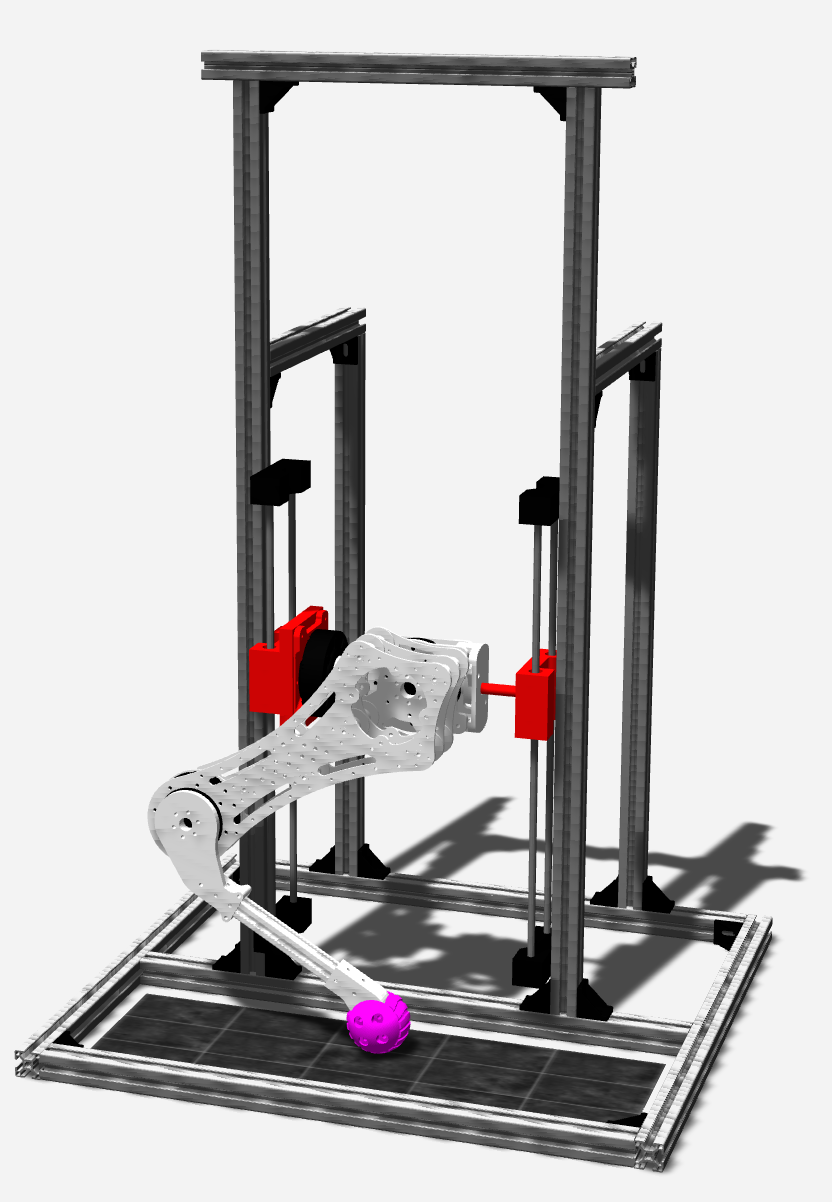}
    \caption{Digital twins of the one-leg stand in Gazebo simulator.
}
    \label{fig:twin}
\end{figure}

\section{Novelty of HyperSurf System}

The HyperSurf system presents several innovative advancements that significantly enhance the capabilities of quadruped robot leg surface recognition, both in real-world and simulated environments. The key novelties of the system can be categorized into three primary areas: boosting the speed and efficiency of dataset collection, optimizing the workforce required for data acquisition, and enhancing the accuracy and transferability of data between physical and simulated environments.

\subsection{Boosting Dataset Collection Efficiency}

One of the most notable innovations of the HyperSurf system is its ability to accelerate the process of dataset collection. Traditionally, collecting datasets for surface recognition involved running extensive tests with a fully operational quadruped robot, which is both time-consuming and resource-intensive. By contrast, HyperSurf introduces a one-leg test setup that significantly streamlines this process. On average, the GRU analysis speed was boosted by a factor of 7.5. This remarkable increase in efficiency is primarily due to the rapid deployment of the system and its capability to operate autonomously in controlled laboratory conditions. The reduction in complexity and the focus on a single leg not only speeds up the data collection process but also allows for more frequent iterations and refinements, leading to a faster development cycle.

\subsection{Optimizing Workforce for Dataset Collection}

The HyperSurf system also demonstrates a significant reduction in the manpower required for dataset collection. In previous approaches, data collection was a labor-intensive process that often required the involvement of more than five team members, particularly due to the extensive duration of the experiments and the need to manage a complex, fully assembled quadruped robot. HyperSurf, however, simplifies the experimental setup by utilizing a static, one-leg configuration in a laboratory environment. This simplification not only shortens the time required for each dataset collection session but also reduces the number of personnel needed. In this study, the entire data collection process was efficiently conducted by just two researchers. This reduction in team size not only lowers operational costs but also minimizes the coordination challenges often associated with larger teams, further streamlining the research and development process.

\begin{figure}
    \centering
    \includegraphics[width=1.0\linewidth]{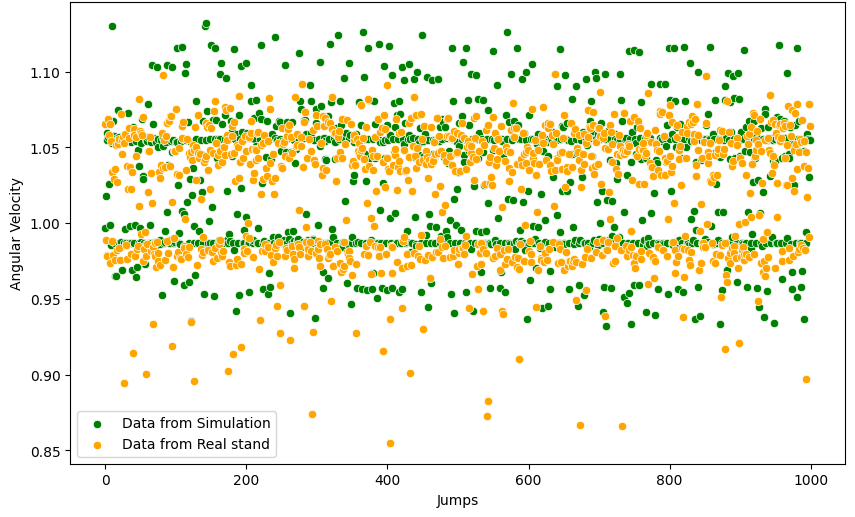}
    \caption{Comparison of the foot angular velocities during the jumps of simulated model and the real robotic leg.}
    \label{fig:comp}
\end{figure}

\subsection{Enhancing Data Transfer and Simulation Accuracy}

A critical advancement introduced by the HyperSurf system is the improved transferability of data between physical and simulated environments. By leveraging data from an IMU sensor placed on the stand, the HyperSurf system enables empirical modeling of surfaces within the Gazebo simulation environment. As shown in Fig. \ref{fig:twin}, this approach facilitates the creation of highly accurate digital twins of physical surfaces. The accuracy of this modeling is highlighted in Fig. \ref{fig:comp}, where a comparison between the angular velocities during jumps of the simulated model and the real robotic leg is presented. The results demonstrate a high degree of correlation, with the simulation achieving an accuracy of 98\% relative to the real-world data.

\section{Conclusion and Future Work}
The HyperSurf technology comprises a specialized leg test setup designed to enhance surface recognition capabilities in quadruped robots. It integrates components such as a mechanical leg with three degrees of freedom, BLDC motors for movement, and an IMU sensor for feedback collection. This setup assists the DogSurf classification model in achieving a precision of up to 98\%. 

Future work will focus on demonstrating how the HyperSurf surface recognition system can enhance quadruped platforms. Feedback from the GRU, processed with the DogSurf architecture, will be utilized to adapt gaits to different surfaces, adjust BLDC motor parameters, and incorporate additional mechanics to improve energy efficiency and terrain traversal. Furthermore, the development of the quadruped platform \cite{HyperSurf.git} will continue, integrating the system to enhance its capabilities based on surface recognition feedback. Additionally, efforts will be made to extend the system to object grasping, leveraging surface property information to improve grasping effectiveness.

In summary, the HyperSurf technology not only marks a significant milestone in surface recognition for quadruped robots but also sets the stage for broader applications across industries such as search and rescue, agriculture, and industrial inspection. This research underscores the potential of integrating advanced sensory and control systems to enable robust and adaptive autonomous systems capable of navigating complex environments with precision and reliability.

\section*{Acknowledgements} 
Research reported in this publication was financially supported by the RSF grant No. 24-41-02039.
Thanks to Liaisan Safarova for installing the operating system on the PC and helping with the assembly!


\bibliographystyle{IEEEtran}
\balance
\bibliography{HyperSurf_Quadruped_Robot_Leg_Capable_of_Surface_Recognition_with_GRU_and_Real-to-Sim_Transferring.bib}


\end{document}